\documentclass{article}

 \usepackage[nonatbib,final]{neurips_2020_tda}

\usepackage[utf8]{inputenc} % allow utf-8 input
\usepackage[T1]{fontenc}    % use 8-bit T1 fonts
\usepackage{amsfonts}       % blackboard math symbols
\usepackage{booktabs}       % professional-quality tables
\usepackage{hyperref}       % hyperlinks
\usepackage{url}            % simple URL typesetting
\usepackage{microtype}      % microtypography
\usepackage{nicefrac}       % compact symbols for 1/2, etc.
\usepackage{paralist}       % in-paragraph enumerations
\usepackage{siunitx}        % SI units

\usepackage{caption}
\usepackage{subcaption}
\usepackage[pdftex]{graphicx}
\usepackage{xcolor}
\usepackage{algorithm}
\usepackage{algorithmic}

\usepackage{booktabs}
\newcommand{\RR}{\mathbb{R}}
\newcommand{\PP}{\text{Pr}}

\usepackage{xcolor}
\usepackage{tikz}
\usepackage{enumitem}
\usetikzlibrary{cd, calc, shapes, decorations.shapes, decorations.markings, decorations.pathreplacing, shapes.geometric, shapes.symbols,positioning}

\tikzset{twosimp/.style={fill opacity=0.6,fill=gray,draw opacity=0.9}}

\title{$k$-simplex2vec: a simplicial extension of node2vec}

\author{%
  Celia Hacker\\
  Departement of Mathematics\\
 EPFL\\
  Lausanne, Switzerland \\
  \texttt{celia.hacker@epfl.ch} \\
}

\begin{document}

\maketitle

\begin{abstract}
We present a novel method\footnote{The code is available at \url{https://github.com/celiahacker/k-simplex2vec}} of associating Euclidean features to simplicial complexes, providing a way to use them as input to statistical and machine learning tools. This method extends the node2vec algorithm to simplices of higher dimensions, providing insight into the structure of a simplicial complex, or into the higher-order interactions in a graph. 
\end{abstract}

\section{Introduction} 
In many areas of research, data comes with its own intrinsic structure. That data might be non-Euclidean, leading in some cases to difficulties in using familiar tools from statistics and machine learning. %If the data is to be represented in a Euclidean space for the purpose of leveraging such tools, 
To address this problem, some methods provide ways to learn how to understand this data in a Euclidean space. It is important to impose some constraints on the Euclidean representation of the data to best render %\celia{recover, respect }
 its initial structure. In other words, we look for a function $F: X \rightarrow\RR^d$ that preserves some of the structure encoded in $X$ and where $d$ is small relative to the size of $X$. One may hope that the image of $F$ retains enough of the original structure of $X$ that it can be discerned using statistical or machine learning tools. This structure-preserving map is often called an \emph{embedding}, and $\RR^d$ the \emph{feature space}. The well known word2vec \cite{word2vecarxiv,word2vec}, and node2vec \cite{node2vec} algorithms are examples of such %embedding 
methods, the former preserving some form of structure in a text, and the latter preserving that of a graph. The idea behind node2vec, which is itself based on word2vec, is to do random walks starting at each vertex of the graph, recording the co-occurrences of nodes in random walks. One then maximizes a function of $\langle F(u),F(v)\rangle$ for vertices $u$, $v$ that appear %more 
%often 
 together in the same random walk, where $\langle \cdot ,\cdot \rangle$ denotes the scalar product in the feature space. 

The aim of this paper is to generalize node2vec to a wider class of objects, namely to \emph{simplicial complexes}. These objects are a higher dimensional analog of graphs, which in addition to nodes ($0$-simplices) and edges ($1$-simplices), also contain triangles ($2$-simplices) and $k$-simplices, which are formed by $k+1$ vertices. In the world of topological data analysis these naturally encode higher-order relations in data and posses a more complex topological structure than graphs \cite{tda}. As a consequence, more structure is taken into consideration while defining random walks on simplicial complexes.  % \celia{Can this sentence be made more positive? Yes, what you say is true, but this also means that the world of these walks is far richer.} 

The first step towards generalizing node2vec is to define a notion of random walk on simplicial complexes. In the graph case the walker can go from one vertex to another whenever they are connected by an edge. For simplices of higher dimension, there are different possibilities of generalization. As stated in \cite{Ori}, a direct generalization of the walk on a graph can be made by connecting any two $k$-simplices if they are a face of the same $(k+1)$-simplex. However, the generalization of the graph Laplacian to simplicial complexes, in \cite{Sayan}, encodes connections in terms of both upper connections through cofaces and lower connections through faces. In what follows we use walks inspired by \cite{Sayan}, using both %upper and lower 
connections. The relationship between random walks and simplicial Laplacians, although not as straightforward as in the graph case \cite{chung}, reveals interesting connections to the algebraic topology of the simplicial complex, as shown in \cite{Ori, Sayan}. 

A generalization of node2vec was proposed in \cite{petri}, in terms of random walks on the Hasse diagram of a simplicial complex, which is a graph encoding face relations. The method then simply applies node2vec to that graph. We believe this approach does not have a strong link to the algebraic topological properties of the simplicial complex, by the nature of the walks being used. In \cite{petri}, only the effect of the algorithm on the nodes is studied, omitting the representations of higher dimensional simplices present in the complex. Other examples of applications of random walks on simplicial complexes can be found in \cite{Schaub}. 

\section{Proposed Method}\label{sec:exp}
A simplicial complex $X$ is a collection of finite sets closed under taking subsets. The building blocks are $k$-simplices, which are sets of cardinality $k+1$. We denote the subset of $k$-simplices of $X$ by $X_k$. The local structure of the simplicial complex is encoded in the two types of connections %, upper and lower, 
 between simplices. Walking on the simplices through upper and lower connections gives a way to transcribe the local structures of the simplicial complex into point clouds using the process described below and in Algortihm \ref{alg:s2v}. The behaviour of the random walks on the $k$-simplices is encoded in a transition matrix $P$, which is a square matrix of size $|X_k|$, with each row summing to $1$. The row $P_\sigma$, corresponding to a simplex $\sigma$, encodes the probability that a walker at $\sigma$ has of jumping to any $k$-simplex in the complex. As mentioned in the introduction, many choices can be made when constructing $P$, yielding many variations of the random walks. %For the experiments in this paper  
Here, we choose to assign equal probability to each neighbor of a simplex, independently of whether it is an upper or lower neighbor. %Many variations of the random walks are possible.  
For this algorithm we perform $N$ random walks starting at each simplex $\sigma\in X_k$, which we truncate at a fixed length $l$. We wish  to obtain a representation $F:X_k \rightarrow \RR^d$ mapping the $k$-simplices into a space of dimension $d$, usually small relative to the size of $X_k$. To obtain this mapping $F$, we move $F(\sigma)$ and $F(\tau)$ closer whenever $\sigma, \tau$ often appear in the same random walks, while on the contrary moving them further apart if they do not appear together in the walks. This optimization is translated into a maximum likelihood problem, using the random walks as similarity measure \cite{deepwalk}. 

We want to maximize the probability of observing a certain random walk starting at a simplex $\sigma$ conditioned on its representation, over possible mappings $F: X_k \rightarrow \RR^d$, i.e., $$\max_F \sum_{\sigma\in X_k} \log(\PP(\text{RW}(\sigma) \vert F(\sigma))),$$ where $\text{RW}(\sigma)$ are the simplices reached by a random walk at $\sigma$. Assuming independence and symmetry in the feature space, the conditional probability is $$\PP (\text{RW}(\sigma)\vert F(\sigma)) = \prod_{\tau \in \text{RW}(\sigma)} \PP (\tau\vert F(\sigma)),$$ where $\PP (\tau \vert F(\sigma)) = \frac{\exp \langle F(\tau), F(\sigma) \rangle}{ \sum_{\nu\in X_k} \exp   \langle F(\nu), F(\sigma) \rangle}. $ This leads to maximizing the following objective function $$\max_F \sum_{\sigma \in X_k}\left[ -\log \left( \sum_{\nu\in X_k}  \langle F(\sigma), F(\nu)\rangle \right) +  \sum_{\tau\in \text{RW}(\sigma)} \langle F(\tau),F(\sigma)\rangle  \right]$$ using stochastic gradient descent. For more details we refer the reader to \cite{node2vec, deepwalk}. We point out that when $k=0$, we recover the node2vec algorithm.
%%%%%%%%%%%Pseudo Algorithm %%%%%%%%%%
\begin{algorithm}
%%%% writing pseudo algorithm 
\caption{$k$-simplex2vec}
\label{alg:s2v}
\begin{algorithmic} 
\REQUIRE(simplicial complex $X$,  dimension of simplices $k$, dimension of feature space $d$, walks per simplex $N$, walk length $l$)
\STATE $X_k = $ $k$-simplices of $X$
\STATE $P$ = matrix of transition probabilities 
\STATE Walks = $[$ $]$
\FOR{i = $1$ to $N$}
\FOR{all $\sigma \in X_k$}
\STATE walk = \textbf{SimplicialWalks}($\sigma$, $P$, $X_k$)
\STATE append walk to Walks
\ENDFOR
\ENDFOR
\STATE	 $F$ = StochasticGradientDescent($X_k$, $d$, Walks)
\RETURN $F$
\\\hrulefill
\end{algorithmic}
\begin{algorithmic}
\ENSURE(start simplex $\sigma$, length $l$, transition matrix $P$)
\STATE  walk = $[\sigma]$
\FOR{$j=1$ to $l$}
\STATE choose $\tau$ from neighbours($\sigma$) according to $P_\sigma$
\STATE append $\tau$ to walk
\ENDFOR
\RETURN walk 
\end{algorithmic}
\end{algorithm}
 
%%%%%%%%%%%%%%%%%%%%%%%%%%%%%%%%%%%%%%%%%%%
%\celia{Computational cost?}
%\todo{Emphasize more the choice of walks!!}
\section{Experimental results} \label{sec:results}
In this section we provide experimental results for the algorithm described above. We have explored several data sets yielding interesting results, however in this paper we choose to focus on one particularly well-understood model of synthetic data, in order to capture the essence of the method. 

\begin{figure}
\centering
\includegraphics[scale=0.2]{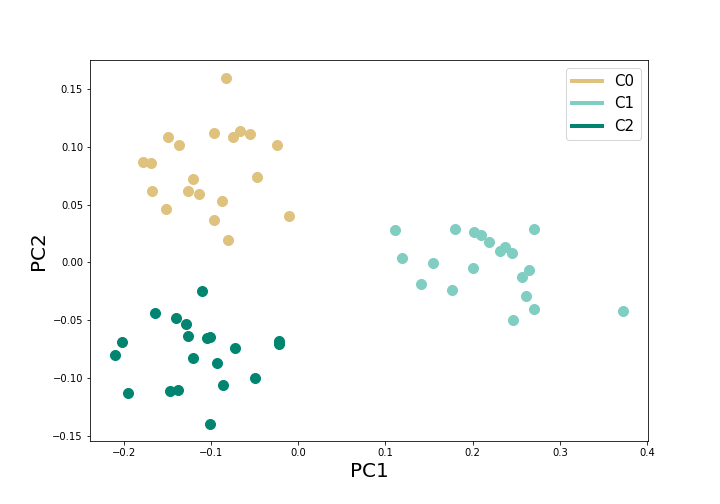}
\caption{The PCA projection of the node2vec representation of the stochastic block model shows three cluster corresponding to the three groups of nodes.}\label{fig:node2vec}
\end{figure}

\textbf{Methodology.} We run the algorithm above ten times on a simplicial complex built on a stochastic block model (SBM) \cite{SBM}. The model consists of three blocks of $20$ vertices, with intra-block probability $0.8$ and inter-block probability $0.3$. We build a simplicial complex out of the SBM by taking its clique complex, where every complete subgraph on $k+1$ vertices gives rise to a $k$-simplex. The complex contains $60$ nodes, $809$ edges and $3610$ triangles. We run the algorithm above for $k=1,2$, i.e., on the edges and the triangles of the clique complex, yielding ten point clouds representing the $1$-simplices and ten point clouds representing the $2$-simplices. On the nodes of the graph, the algorithm is equivalent to node2vec. In this case the result is discussed in \cite{sbm_node2vec} and is shown in Figure \ref{fig:node2vec}, where the clusters perfectly recover the three groups of vertices of the original SBM. %We extend this result to higher dimensional simplices in the following way. 
 This result can be extended in the following way to simplices of higher dimension. 

The $1$-simplices and $2$-simplices can be sorted into different classes. There are two types of $1$-simplices, those that have both nodes within one block and those with nodes in two different blocks, yielding a total of six classes, depending in which blocks the vertices are. In the same way there are three sorts of triangles in the clique complex. We distinguish $2$-simplices that have all vertices inside the same block, $2$-simplices that have two vertices in one block and the third in another, and finally $2$-simplices with exactly one vertex in each block. These yield ten classes of $2$-simplices, depending on the blocks the vertices belong to. The types of simplices are shown in Figures \ref{fig:edges} and \ref{fig:triangles}. 
 
Various parameters of the algorithm can be tuned, i.e., the length of the walks, the number of walks starting at each simplex, and the dimension of the feature space, which is in general much less than $\vert X_k\vert$. We vary these parameters within a certain range. For instance, the number of walks and their length vary between $10$ and $50$ for the edges and between $20$ and $60$ for the $2$-simplices. We consider feature spaces of dimension $10$, $20$, and $30$. After running the algorithm on the edges and triangles, we apply a clustering algorithm to the resulting point clouds and compare the clusters with the known types of $k$-simplices using the Rand index \cite{rand}. The Rand index compares two label assignments to the same set of points, measuring the proportion of points with the same labels in both cases.
%%%%%%%%%%%%%%%%%%%%%%%%%%%%

\begin{figure*}[h]%[t!]
\centering
    %%%%%%%%%%%%%%%
    \begin{subfigure}[b]{0.5\linewidth}
    \centering
        \includegraphics[width=0.77\textwidth]{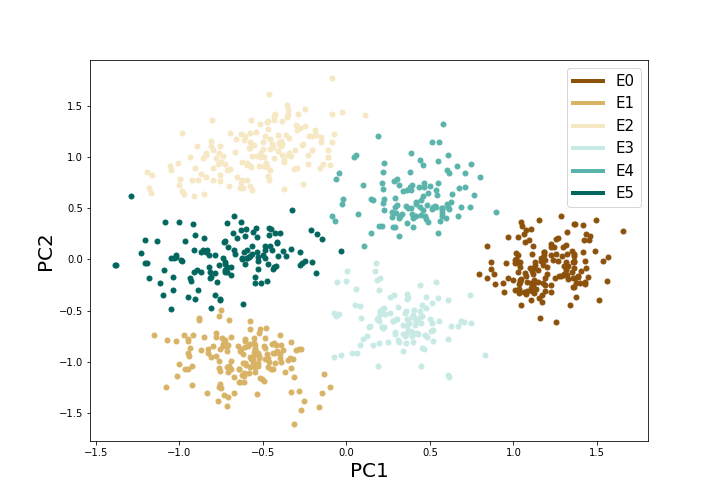}
        \caption{Clusters of edges from the feature space of dimension $20$, using a walk length of $20$ and $40$ walks per edge.\vspace{10 pt }} \label{fig:cluster_edges}
    \end{subfigure}%
    ~ 
    %%%%%%%%%%%%%%%%%%%%%%%%%%%
    \begin{subfigure}[b]{0.5\linewidth}
\centering
        \includegraphics[width=0.77\textwidth]{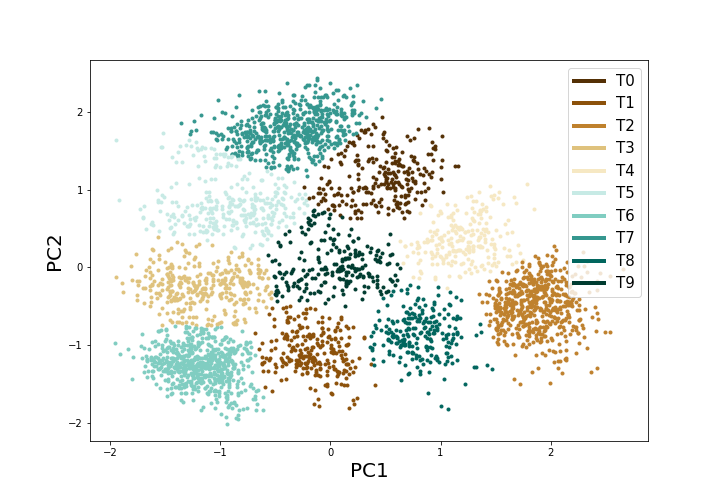}
        \caption{Clusters of triangles from the feature space of dimension $20$, using a walk length of $20$ and $40$ walks per triangle.}\label{fig:cluster_triangles}
    \end{subfigure}
    %%%%%%%%%%%%%%%%%%%%%%%
  \caption{Both figures were obtained from the features representing the $1$ and $2$-simplices by applying PCA. The colors reflect the clusters determined by the kmeans algorithm.}\label{fig:PCA}
\end{figure*}

\textbf{Results.} We perform clustering on the point clouds using the kmeans algorithm. The results of the clustering for the edges and triangles are plotted in Figure \ref{fig:PCA} using PCA in two dimensions for a given set of parameters. We run the clustering with $k= 6$ for the edges and $k=10$ for the $2$-simplices to reflect the types of $1$ and $2$- simplices present in the clique complex of the SBM. We study the clusters obtained on the point clouds by comparing them to the ground truth given by the types of simplices discussed above. The resulting Rand indices over ten trials of the algorithm are presented in Tables \ref{Tab:rand_edges} and \ref{Tab:rand_triangles}, for a fixed walk length of $20$. We consider feature spaces of dimensions $10$ , $20$, $30$ and vary the number of walks starting at each simplex. Results using the DBSCAN algorithm for clustering instead of kmeans can be found in Section \ref{sec:sup_res}.  %\celia{reference to section}

The six classes of edges are recovered with an accuracy above $0.97$, and the ten classes of triangles with a Rand index above $0.80$, reaching $0.88$ in the feature space of dimension $30$. As we can see in this range of parameters, varying the number of walks for a fixed length does not change the accuracy. Rather, the dimension of the feature space influences the accuracy of the clustering. In this range of parameters the higher the dimension, the greater the precision. As illustrated in these results, this method allows us to distinguish, with high precision, among different classes of simplices in the clique complex of a graph, directly generalizing the result for node2vec on the vertices. Through these results, we gain insight into the higher dimensional structure of the underlying graph, highlighting the many types of interactions between groups of vertices present in the graph.

 %%%%%% Table for graph stats 

\begin{table}
\begin{center}
%\scriptsize
\caption{Rand index for varying number of walks and dimension of the feature space of edges. }\label{Tab:rand_edges}
 \begin{tabular}{c c c c c c} 
\toprule
 & & & Number of walks  & & \\
dimensions & 10 & 20 &30 & 40 & 50  \\ \hline

 $10$ & $0.98\pm 0.01$ & $0.98\pm 0.00$ & $0.98\pm 0.01$ & $0.98\pm 0.01$ & $0.98\pm 0.01$ \\ 

  $20$ & $0.98\pm 0.01$ & $0.98\pm 0.01$ & $0.99\pm 0.00$ & $0.98\pm 0.00$ & $0.99\pm 0.00$ \\ 

  $30$ & $0.99\pm 0.01$ & $0.99\pm 0.01$ & $0.99\pm 0.01$ &$0.99\pm 0.04 $ & $0.99\pm 0.00$  \\ 
\bottomrule
\end{tabular}

\caption{Rand index for varying number of walks and dimension of the feature space of triangles.}\label{Tab:rand_triangles}
 \begin{tabular}{c  c c c c c} 
\toprule
 & & & Number of walks & & \\
dimensions  & 20 &30 & 40 & 50 & 60 \\\hline

 $10$ & $0.81\pm 0.10$ & $0.82\pm 0.10$ & $0.81\pm 0.10 $ & $0.82\pm 0.10$ & $0.82\pm 0.10$ \\ 

  $20$ & $0.88\pm 0.10$ & $0.88\pm 0.10$ & $0.88\pm 0.10$ & $0.88\pm 0.10$ & $0.88\pm 0.10$ \\ 

  $30$ & $0.88\pm 0.10$ & $0.88\pm 0.10$ & $0.88\pm 0.10$ & $0.88\pm 0.10$ & $0.88\pm 0.10$ \\ 
\bottomrule
\end{tabular}
\end{center}
\end{table}

\section{Conclusion} 
The results presented here are promising. We have obtained similiar results for different types of network data, such as other random graph models, social networks and neural data. For the sake of brevity and clarity of the message, these were not included in the paper.  Understanding the results on a graph with well known structure is key to understanding how the algorithm works and provides intuition for more complicated data sets. 

So far we have considered the algorithm only on simplicial complexes obtained from network structured data. We seek to extend the use of this method to simplicial complexes that are not necessarily clique complexes of graphs. Further work includes a thorough study of the effect of changing each hyperparameter of the algorithm, as well as studying the effect of different types of walks, as mentioned in Section \ref{sec:exp}. After further investigation of the results using this algorithm we will study the algorithmic aspects such as the complexity and performance of the method.

In analogy to the graph case where the random walks detect the connected components, i.e., corresponding to homology in dimension $0$, the random walks on simplicial complexes as described in \cite{Sayan} and \cite{Ori} have an interesting connection to global algebraic topology properties of the simplicial complex. We believe that with the method described in this paper, we can extract other topologically interesting features by means of random walks on simplicial complexes. 

\begin{ack} I thank Kathryn Hess for guidance and advice throughout this project, as well as Gard Spreemann and Stefania Ebli for the useful discussions. The author was supported by the NCCR Synapsy grant of the Swiss National Science Foundation.
\end{ack}

%\section*{References}
\bibliography{NeurIPS_refs}
\bibliographystyle{plain}  
%\newpage
%\clearpage
\appendix
\section{Supplementary Materials}

\subsection{Random walks on simplicial complexes}
In this section we illustrate the upper and lower connections that appear in simplicial complexes. As in Section \ref{sec:exp}, one can connect two simplices of same dimension by a connection through either a common face or a common coface. In either case the difference in dimension is $1$. 

\begin{figure}[H]
\begin{center}
\begin{tikzpicture}
\input{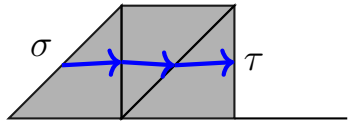}
\end{tikzpicture}
\caption{A walk from $\sigma$ to $\tau$ using only upper connections, i.e., through common triangles. }\label{fig:upper}
\end{center}
\end{figure}

\begin{figure}[H]
\begin{center}
\begin{tikzpicture}
\input{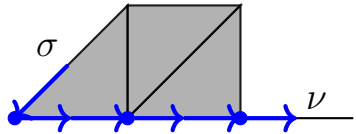}
\end{tikzpicture}
\caption{A walk from $\sigma$ to $\nu$ using only lower connections, namely through vertices.}\label{fig:lower}
\end{center}
\end{figure}

Figure \ref{fig:upper} illustrates the walks using upper connections only: a walk from one edge to another, through common triangles. In contrast, Figure \ref{fig:lower} shows that simplex $\sigma$ and $\nu$ are connected through a walk using only connections through shared $0$-simplices. One can consider these two types of walks separately, or use them together, mixing both types, as is done in Algorithm \ref{alg:s2v}.

\subsection{Results} \label{sec:sup_res}
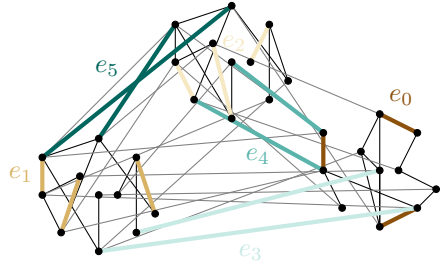
\begin{figure}
\begin{center}
   \begin{tikzpicture}
 %%% Picture for SBM edges

\begin{scope}[scale = 0.25]

\definecolor{c1}{RGB}{140,81,10}
\definecolor{c2}{RGB}{216,179,101}
 \definecolor{c3}{RGB}{ 246,232,195}
 \definecolor{c4}{RGB}{199,234,229}
 \definecolor{c5}{RGB}{90,180,172}
 \definecolor{c6}{RGB}{1,102,94}

 %%% Block 1 
\coordinate (x1) at (1,4);
\coordinate(x2) at (2,2);
\coordinate (x3) at (4,1); 
\coordinate (x4) at (1,6); 
\coordinate (x5) at  (3,5); 
\coordinate (x6) at  (4,4); 
\coordinate (x7) at  (6,2); 
\coordinate (x8) at  (4,7); 
\coordinate (x9) at  (5,4); 
\coordinate (x10) at  (7,3); 
\coordinate (x11) at  (6,6); 

  %%% Block 2
 \coordinate (y1) at ($(x1) + (45:10)$);
\coordinate(y2) at ($(x2) + (45:10)$);
\coordinate (y3) at   ($(x3) + (45:10)$);
\coordinate (y4) at  ($(x4) + (45:10)$);
\coordinate (y5) at   ($(x5) + (45:10)$);
\coordinate (y6) at   ($(x6) + (45:10)$);
\coordinate (y7) at   ($(x7) + (45:10)$);
\coordinate (y8) at   ($(x8) + (45:10)$);
\coordinate (y9) at   ($(x9) + (45:10)$);
\coordinate (y10) at   ($(x10) + (45:10)$);
\coordinate (y11) at   ($(x11) + (45:10)$);

 %%% Block 3
 \coordinate (z1) at ($(x1) + (5:15)$);
\coordinate(z2) at  ($(x2) +(5:15)$);
\coordinate (z3) at   ($(x3) +(5:15)$);
\coordinate (z4) at  ($(x4) + (5:15)$);
\coordinate (z5) at   ($(x5) + (5:15)$);
\coordinate (z6) at   ($(x6) + (5:15)$);
\coordinate (z7) at   ($(x7) + (5:15)$);
\coordinate (z8) at   ($(x8) +(5:15)$);
\coordinate (z9) at   ($(x9) + (5:15)$);
\coordinate (z10) at   ($(x10) + (5:15)$);
\coordinate (z11) at   ($(x11) + (5:15)$);

%% Edges in block 1  
\draw[ black] (x1) --(x2);
\draw[black] (x1) --(x4);
\draw[ black] (x3) --(x6);
\draw[black] (x4) --(x3);
\draw[ black] (x1) --(x5);
\draw[ black] (x7) --(x11);
\draw[ black] (x10) --(x11);
\draw[ black] (x8) --(x10);
\draw[ black] (x9) --(x11);
\draw[ black] (x8) --(x4);
 \draw[ black] (x8) --(x2);
 
 %% Edges in block 2 
\draw[ black] (y1) --(y2);
\draw[ black] (y1) --(y4);
\draw[ black] (y3) --(y6);
\draw[ black] (y4) --(y3);
\draw[ black] (y1) --(y5);
\draw[ black] (y7) --(y11);
\draw[black] (y10) --(y11);
\draw[black] (y8) --(y10);
\draw[ black] (y9) --(y11);
\draw[, black] (y8) --(y4);
 \draw[ black] (y8) --(y2);

   %% Edges in block 3
\draw[black] (z1) --(z2);
\draw[ black] (z1) --(z4);
\draw[black] (z3) --(z6);
\draw[black] (z5) --(z3);
\draw[ black] (z6) --(z5);
\draw[ black] (z7) --(z1);
\draw[ black] (z10) --(z7);
\draw[ black] (z8) --(z5);
\draw[ black] (z9) --(z11);
\draw[black] (z8) --(z3);
\draw[black] (z9) --(z10);
  
  %% Edges between 1 and 3 
\draw[ gray] (z1) --(x2);
\draw[gray] (z4) --(x4);
\draw[gray] (z10) --(x6);
\draw[ gray] (z5) --(x3);
\draw[ gray] (z6) --(x5);
\draw[ gray] (z7) --(x1);

  %% Edges between 1 and 3 
\draw[ gray] (y1) --(x2);
\draw[gray] (y4) --(x4);
\draw[ gray] (y10) --(x3);
\draw[ gray] (y5) --(x8);
\draw[ gray] (y6) --(x9);
\draw[ gray] (y7) --(x1);

   %% Edges between 1 and 3 
\draw[ gray] (y1) --(z2);
\draw[gray] (y4) --(z4);
\draw[ gray] (y8) --(z3);
\draw[ gray] (y5) --(z8);
\draw[ gray] (y2) --(z9);
\draw[gray] (y3) --(z1);

 %% draw  edges of clustering 
 
%  \draw[ultra thick, c1] (x4) --(x10);
\draw[ ultra thick, c2] (x1) --(x4);
 \draw[ultra thick, c2] (x5) --(x2);
  \draw[ultra thick, c2] (x10) --(x11);
  
  \draw[ultra thick, c3] (y9) --(y11);
\draw[ultra  thick, c3] (y1) --(y2);
 \draw[ ultra thick, c3] (y3) --(y5);
 
% \draw[ultra thick, c3] (z7) --(z6);
% \draw[ ultra thick, c3] (z9) --(z2);
\draw[ultra thick, c1] (z4) --(z1);
\draw[ultra thick, c1] (z3) --(z7);
 \draw[ultra thick, c1] (z8) --(z11);
 
% \draw[ultra thick, c4] (z4) --(x3);
\draw[ultra thick, c4] (z6) --(x7);
\draw[ultra thick, c4] (z7) --(x3);

% \draw[ultra thick, c5] (y5) --(x2);
\draw[ultra thick, c6] (y4) --(x8);
\draw[ultra thick,c6] (y8) --(x4);

%\draw[ultra thick, c6] (y4) --(z2);
\draw[ultra thick, c5] (y6) --(z4);
\draw[ultra thick, c5] (y2) --(z1);

\draw[black, fill = black] (x1) circle (5pt);
\draw[black, fill = black] (x2) circle (5pt);
\draw[black, fill = black] (x3) circle (5pt);
\draw[black, fill = black] (x4) circle (5pt);
\draw[black, fill = black] (x5) circle (5pt);
\draw[black, fill = black] (x6) circle (5pt);
\draw[black, fill = black] (x7) circle (5pt);
\draw[black, fill = black] (x8) circle (5pt);
\draw[black, fill = black] (x9) circle (5pt);
\draw[black, fill = black] (x10) circle (5pt);
\draw[black, fill = black] (x11) circle (5pt);

\draw[black, fill = black] (y1) circle (5pt);
\draw[black, fill = black] (y2) circle (5pt);
\draw[black, fill = black] (y3) circle (5pt);
\draw[black, fill = black] (y4) circle (5pt);
\draw[black, fill = black] (y5) circle (5pt);
\draw[black, fill = black] (y6) circle (5pt);
\draw[black, fill = black] (y7) circle (5pt);
\draw[black, fill = black] (y8) circle (5pt);
\draw[black, fill = black] (y9) circle (5pt);
\draw[black, fill = black] (y10) circle (5pt);
\draw[black, fill = black] (y11) circle (5pt);

\draw[black, fill = black] (z1) circle (5pt);
\draw[black, fill = black] (z2) circle (5pt);
\draw[black, fill = black] (z3) circle (5pt);
\draw[black, fill = black] (z4) circle (5pt);
\draw[black, fill = black] (z5) circle (5pt);
\draw[black, fill = black] (z6) circle (5pt);
\draw[black, fill = black] (z7) circle (5pt);
\draw[black, fill = black] (z8) circle (5pt);
\draw[black, fill = black] (z9) circle (5pt);
\draw[black, fill = black] (z10) circle (5pt);
\draw[black, fill = black] (z11) circle (5pt);

%%%%%% Edge names 
%%%% e0
%\coordinate(e0) at  (y9);
\node[anchor = south west] at (z8) {\color{c1} $e_0$};
%%%%% e1
\node[anchor = north east] at (x4) {\color{c2} $e_1$};
%%%%e2
\coordinate(e2) at ($(y9)+(0,0.5)$);
\node[anchor = west] at (y5) {\color{c3} $e_2$};
%%%e3
\coordinate(e3) at   ($(x3) + (7:7)$);
\node[anchor = north west] at (e3) {\color{c4} $e_3$};

%%%e4
\coordinate(e4) at   ($(y2) - (155:5)$);
\node[anchor = north east] at (e4) {\color{c5} $e_4$};

%%%e3
\coordinate(e5) at   ($(x4) + (40:6)$);
\node[anchor = south east] at (e5) {\color{c6} $e_5$};

\end{scope}

  
 \end{tikzpicture}
 \caption{The edges are colored corresponding to the clusters they belong to in Figure \ref{fig:cluster_edges}, namely an edge $e_i$ corresponds to a point in cluster $E_i$.}\label{fig:edges}
\end{center}
\end{figure}

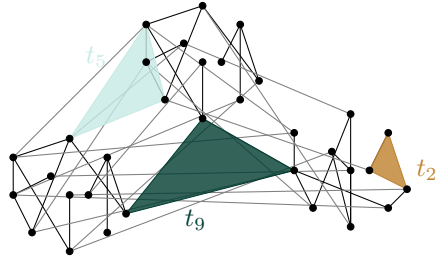
\begin{figure}
\begin{center}
   \begin{tikzpicture}
 %% SBM triangles 

\definecolor{d1}{RGB}{84,48,5}
\definecolor{d2}{RGB}{140,81,10}
 \definecolor{d3}{RGB}{ 191,129,45}
 \definecolor{d4}{RGB}{223,194,125}
 \definecolor{d5}{RGB}{246,232,195}
 \definecolor{d6}{RGB}{199,234,229}
\definecolor{d7}{RGB}{128,205,193}
 \definecolor{8}{RGB}{53,151,143}
 \definecolor{d9}{RGB}{1,102,94}
 \definecolor{d10}{RGB}{0,60,48}

\begin{scope}[scale = 0.25]
 %%% Block 1 
\coordinate (x1) at (1,4);
\coordinate(x2) at (2,2);
\coordinate (x3) at (4,1); 
\coordinate (x4) at (1,6); 
\coordinate (x5) at  (3,5); 
\coordinate (x6) at  (4,4); 
\coordinate (x7) at  (6,2); 
\coordinate (x8) at  (4,7); 
\coordinate (x9) at  (5,4); 
\coordinate (x10) at  (7,3); 
\coordinate (x11) at  (6,6); 
  %%% Block 2
 \coordinate (y1) at ($(x1) + (45:10)$);
\coordinate(y2) at ($(x2) + (45:10)$);
\coordinate (y3) at   ($(x3) + (45:10)$);
\coordinate (y4) at  ($(x4) + (45:10)$);
\coordinate (y5) at   ($(x5) + (45:10)$);
\coordinate (y6) at   ($(x6) + (45:10)$);
\coordinate (y7) at   ($(x7) + (45:10)$);
\coordinate (y8) at   ($(x8) + (45:10)$);
\coordinate (y9) at   ($(x9) + (45:10)$);
\coordinate (y10) at   ($(x10) + (45:10)$);
\coordinate (y11) at   ($(x11) + (45:10)$);

 %%% Block 3
 \coordinate (z1) at ($(x1) + (5:15)$);
\coordinate(z2) at  ($(x2) +(5:15)$);
\coordinate (z3) at   ($(x3) +(5:15)$);
\coordinate (z4) at  ($(x4) + (5:15)$);
\coordinate (z5) at   ($(x5) + (5:15)$);
\coordinate (z6) at   ($(x6) + (5:15)$);
\coordinate (z7) at   ($(x7) + (5:15)$);
\coordinate (z8) at   ($(x8) +(5:15)$);
\coordinate (z9) at   ($(x9) + (5:15)$);
\coordinate (z10) at   ($(x10) + (5:15)$);
\coordinate (z11) at   ($(x11) + (5:15)$);

%% Edges in block 1  
\draw[ black] (x1) --(x2);
\draw[black] (x1) --(x4);
\draw[ black] (x3) --(x6);
\draw[black] (x4) --(x3);
\draw[ black] (x1) --(x5);
\draw[ black] (x7) --(x11);
\draw[ black] (x10) --(x11);
\draw[ black] (x8) --(x10);
\draw[ black] (x9) --(x11);
\draw[ black] (x8) --(x4);
 \draw[ black] (x8) --(x2);
 %% Edges in block 2 
\draw[ black] (y1) --(y2);
\draw[ black] (y1) --(y4);
\draw[ black] (y3) --(y6);
\draw[ black] (y4) --(y3);
\draw[ black] (y1) --(y5);
\draw[ black] (y7) --(y11);
\draw[black] (y10) --(y11);
\draw[black] (y8) --(y10);
\draw[ black] (y9) --(y11);
\draw[, black] (y8) --(y4);
 \draw[ black] (y8) --(y2);
   %% Edges in block 3
\draw[black] (z1) --(z2);
\draw[ black] (z1) --(z4);
\draw[black] (z3) --(z6);
\draw[black] (z5) --(z3);
\draw[ black] (z6) --(z5);
\draw[ black] (z7) --(z1);
\draw[ black] (z10) --(z7);
\draw[ black] (z8) --(z2);
\draw[ black] (z9) --(z11);
\draw[black] (z8) --(z3);
  %% Edges between 1 and 3 
\draw[ gray] (z1) --(x2);
\draw[gray] (z4) --(x4);
\draw[gray] (z10) --(x6);
\draw[ gray] (z5) --(x3);
\draw[ gray] (z6) --(x5);
\draw[ gray] (z7) --(x1);
  %% Edges between 1 and 3 
\draw[ gray] (y1) --(x2);
\draw[gray] (y4) --(x4);
\draw[ gray] (y10) --(x3);
\draw[ gray] (y5) --(x8);
\draw[ gray] (y6) --(x9);
\draw[ gray] (y7) --(x1); 
   %% Edges between 1 and 3 
\draw[ gray] (y1) --(z2);
\draw[gray] (y4) --(z4);
\draw[ gray] (y8) --(z3);
\draw[ gray] (y5) --(z8);
\draw[ gray] (y2) --(z9);
\draw[gray] (y3) --(z1);

%%%  triangles 

\tikzset{twosimp/.style={fill opacity=0.8,draw opacity=1}} 

 \draw[twosimp, d3, fill  = d3] (z9) -- (z10) -- (z11) -- cycle;

 \draw[twosimp, d6, fill  = d6] (x8) -- (y4) -- (y2) -- cycle;

\draw[twosimp, d10, fill  = d10] (x10) -- (y3) -- (z1) -- cycle;

\draw[black, fill = black] (x1) circle (5pt);
\draw[black, fill = black] (x2) circle (5pt);
\draw[black, fill = black] (x3) circle (5pt);
\draw[black, fill = black] (x4) circle (5pt);
\draw[black, fill = black] (x5) circle (5pt);
\draw[black, fill = black] (x6) circle (5pt);
\draw[black, fill = black] (x7) circle (5pt);
\draw[black, fill = black] (x8) circle (5pt);
\draw[black, fill = black] (x9) circle (5pt);
\draw[black, fill = black] (x10) circle (5pt);
\draw[black, fill = black] (x11) circle (5pt);

\draw[black, fill = black] (y1) circle (5pt);
\draw[black, fill = black] (y2) circle (5pt);
\draw[black, fill = black] (y3) circle (5pt);
\draw[black, fill = black] (y4) circle (5pt);
\draw[black, fill = black] (y5) circle (5pt);
\draw[black, fill = black] (y6) circle (5pt);
\draw[black, fill = black] (y7) circle (5pt);
\draw[black, fill = black] (y8) circle (5pt);
\draw[black, fill = black] (y9) circle (5pt);
\draw[black, fill = black] (y10) circle (5pt);
\draw[black, fill = black] (y11) circle (5pt);

\draw[black, fill = black] (z1) circle (5pt);
\draw[black, fill = black] (z2) circle (5pt);
\draw[black, fill = black] (z3) circle (5pt);
\draw[black, fill = black] (z4) circle (5pt);
\draw[black, fill = black] (z5) circle (5pt);
\draw[black, fill = black] (z6) circle (5pt);
\draw[black, fill = black] (z7) circle (5pt);
\draw[black, fill = black] (z8) circle (5pt);
\draw[black, fill = black] (z9) circle (5pt);
\draw[black, fill = black] (z10) circle (5pt);
\draw[black, fill = black] (z11) circle (5pt);

%%%%%%%Triangles %%%%%%%%%%

%%%%% t9 
\coordinate (t9) at ($(x7) + (20:5)$); 
\node[anchor = north] at (t9) {\color{d10}$t_9$};

%%%%% t2
\node[anchor = south west] at (z10) {\color{d3}$t_2$};

%%%%%%%%%  t5 
\coordinate (t5) at ($(x8) + (60:5)$); 
\node[anchor = east] at (t5) {\color{d6}$t_5$};

\end{scope}
 \end{tikzpicture}
\caption{The three triangles depict some of the types of triangles distinguished by the clusters of Figure \ref{fig:cluster_triangles}. Triangle $t_2$, with all three vertices inside the same block corresponds to a point in cluster $T_2$. Triangle $t_5$, with two vertices in one block and the third in another one, illustrates the points in cluster $T_5$. Finally, triangle $t_9$ illustrates the $2$-simplices which have exactly one vertex in each block, corresponding to points in cluster $T_9$.}\label{fig:triangles}
\end{center}
\end{figure}

In this section we complete the results shown in Section \ref{sec:results}. The first additions are Figures \ref{fig:edges} and \ref{fig:triangles}, both depicting an example of a stochastic block model with three communities. In Figure \ref{fig:edges} edges are highlighted by colors corresponding to the clusters shown in Figure \ref{fig:cluster_edges}, distinguishing between edges within a block or between two blocks. In Figure \ref{fig:triangles}, three types of triangles are shown, corresponding to three of the clusters in Figure \ref{fig:cluster_triangles}, which distinguish the triangles into ten classes. 

In the results of Section \ref{sec:results}, we show only the clustering using kmeans, which relies on some form of knowledge of the data. We present here some results using the DBSCAN algorithm, which clusters points based on density. 

\begin{figure}
\begin{center}
\includegraphics[scale=0.3]{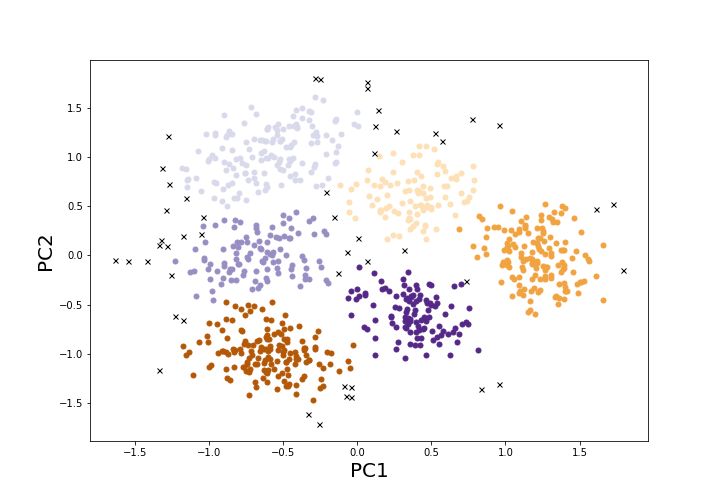}
\caption{The plot shows the PCA in two dimensions of the feature space of dimension $20$, length of walks $10$ and $10$ walks per edge. The points marked by crosses correspond to those that were not clustered by the DBSCAN algorithm. }\label{fig:dbscan_edges}
\end{center}
\end{figure}

Figure \ref{fig:dbscan_edges} shows a clustering on the feature space of the edges, using DBSCAN. The algorithm does not cluster all points, some remain unclustered. The proportion of unclustered points is on average $5\%$ for the feature space of the edges. We discard those points when computing the Rand index for the resulting clusters. Figure \ref{fig:dbscan_edges} shows a clustering using DBSCAN on the feature space of the triangles. In this case, the average proportion of unclustered points is $15\% $.

The Rand indices over ten trials are shown in Tables \ref{Tab:dbscan_edges} and \ref{Tab:dbscan_triangles}. Again, the length of the walks is fixed to be $20$, and we vary the dimension and number of walks just as in the results presented in Section \ref{sec:results}. In this case the Rand indices obtained using the DBSCAN algorithm are approximately the same as those obtained using kmeans. 

\begin{figure}
\begin{center}
\includegraphics[scale=0.3]{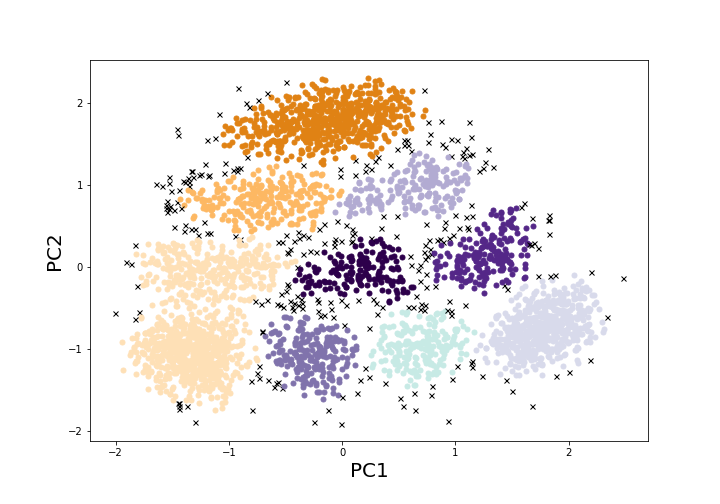}
\caption{The plot shows the PCA in two dimensions of the feature space of dimension $20$ for the edges, with length of walks $10$ and $10$ walks per $2$-simplex. The points marked by crosses correspond to those that were not clustered by the DBSCAN algorithm.}\label{fig:dbscan_triangles}
\end{center}
\end{figure}

\begin{table}
\begin{center}
\caption{Rand index for varying number of walks and dimension of the feature space of edges, with fixed walk length $20$.}\label{Tab:dbscan_edges}
 \begin{tabular}{c c c c c c} 
\toprule
 & & & Number of walks & & \\
dimensions & 10 & 20 &30 & 40 & 50  \\ \hline

 $10$ & $0.98\pm 0.05 $& $ 0.98\pm 0.05 $ & $ 0.99\pm 0.00 $ & $0.98\pm 0.05$ & $ 0.98\pm 0.05$ \\ 

  $20$ & $ 0.96 \pm 0.03$ & $0.96\pm 0.15 $ & $0.98 \pm 0.04$ & $0.99\pm 0.00 $ & $ 0.99 \pm 0.05 $ \\ 

  $30$ & $ 0.89\pm 0.10$ & $ 0.96\pm 0.15$ & $0.95\pm 0.05$ &$ 0.98\pm 0.04 $ & $0.95 \pm 0.15 $  \\ 
\bottomrule
\end{tabular} 
\caption{Rand index for varying number of walks and dimension of the feature space of triangles, with fixed walk length $20$.}\label{Tab:dbscan_triangles}
 \begin{tabular}{c c c c c c} 
\toprule
 & & & Number of walks & & \\
dimensions  & 20 &30 & 40 & 50 & 60  \\\hline

 $10$ & $0.83\pm 0.12$ & $ 0.81\pm 0.15 $ & $ 0.83 \pm 0.15 $ & $ 0.83\pm 0.15$ & $0.81\pm 0.13 $ \\ 

  $20$ & $0.85\pm 0.11 $ & $0.84 \pm 0.11 $ & $0.84\pm 0.12 $ & $ 0.85\pm 0.11$ & $0.85\pm 0.11$ \\ 

  $30$ & $ 0.80\pm 0.12$ & $0.79\pm 0.16$ & $ 0.80\pm 0.12$ & $ 0.80\pm 0.14$ & $ 0.79 \pm 0.15$ \\ 
\bottomrule
\end{tabular}
\end{center}
\end{table}

%\todo{
%Details to be polished
%\begin{itemize} 
%\item which clustering algo (with ref?)
%\item reference tda papers
%\item reference deepwalk \cite{deepwalk} maybe
%\item reference simplicial laplacians 
%\item finish results 
%\item details of SBM  (number of nodes etc) 
%\item is it ok to call it BM? 
%\item cite similarity measure random walks on graphs
%\end{itemize}}

\end{document}